%% file: main.tex
\pdfoutput=1

\documentclass[11pt]{article}
\usepackage{amsfonts}
\usepackage{acl}

\usepackage{pifont}

\usepackage{makecell}
\usepackage{colortbl}
\usepackage[table]{xcolor}
\usepackage{amsmath}
\usepackage{url}
\usepackage{pgfplots}
\pgfplotsset{compat=1.18}

\definecolor{poslight}{RGB}{220,245,220}
\definecolor{posdark}{RGB}{170,225,170}
\definecolor{neglight}{RGB}{245,220,220}
\definecolor{negdark}{RGB}{235,180,180}
\usepackage{array}
\newcolumntype{P}[1]{>{\raggedright\arraybackslash}p{#1}}

\usepackage{times}
\usepackage{latexsym}
\usepackage[normalem]{ulem}

\usepackage{adjustbox}

\usepackage[T1]{fontenc}

\usepackage{booktabs}
\usepackage{subcaption}
\usepackage{tcolorbox}
\usepackage{enumitem}

\usepackage[utf8]{inputenc}

\usepackage{microtype}
\usepackage{multirow}
\usepackage{inconsolata}

\usepackage{graphicx}

\title{LaViSA: A Language and Vision Structural Ambiguity Benchmark}

\author{
Lee Sangmyeong$^{1,2,3}$  
\hspace{1cm}
Shun Inadumi$^{1,2,3}$  
\hspace{1cm}
Koichiro Yoshino$^{3,2,1}$ \\
\\ 
$^1$Nara Institute of Science and Technology 
$^2$Guardian Robot Project RIKEN \\
$^3$The University of Osaka\\
\\ 
\texttt{
lee.sangmyeong.lo3@is.naist.jp
inazumi.shun.in6@naist.ac.jp}\\
\texttt{
yoshino.koichiro.es@osaka-u.ac.jp}
}

\begin{document}
\maketitle

\input{contents/0_abstract}
\input{contents/1_introduction}
\input{contents/2_related_work}

\input{contents/3_LaViSA}
\input{contents/4_experiment}
\input{contents/5_conclusion}

\section*{Limitations}

LaViSA relies entirely on artificially synthesised visual scenes in order to construct controlled image-disambiguated sentence pairs for visual disambiguation.
While synthetic image generation enables efficient construction of examples with diverse semantic contexts and controllable visual content, it may also introduce biases associated with particular image styles or image generation models.
Evaluation using naturally collected images, therefore, remains an important direction for future work.
To mitigate this concern, we conducted human validation across LaViSA and showed that the generated images generally convey the intended interpretations of the corresponding disambiguated sentences.


Our qualitative analysis is based on rationales generated by a single reasoning-oriented model family and focuses on representative failure cases. 
While the analysis provides insight into possible difficulties of VLMs in grounding predicate--argument structures from visual scenes, these observations may not generalise to all VLM architectures or reasoning mechanisms. 
Further investigation is needed to better understand how VLMs internally represent structurally grounded semantics. 


\section*{Acknowledgements}

A part of this work was supported by JST PRESTO grant number JPMJPR24TC. This work was also supported by Guardian Robot Project RIKEN and NAIST Granite Program. 

\bibliography{custom}
\appendix
\input{contents/z_appendix}

\end{document}

%% file: contents/0_abstract.tex
\begin{abstract}

Structural ambiguity arises when a single sentence admits multiple valid interpretations due to its syntactic structure, posing a fundamental challenge for language understanding. 
Visual scenes serve as useful cues for resolving such ambiguity, and Vision and Language Models (VLMs) need to be capable of deriving possible semantic interpretations from visual scenes. We introduce \textbf{La}nguage and \textbf{Vi}sion \textbf{S}tructural \textbf{A}mbiguity (LaViSA), a benchmark designed to evaluate the ability of VLMs to resolve structural ambiguity leveraging visual scenes. 
LaViSA consists of ambiguous sentences, their disambiguated sentences, and corresponding images of these disambiguated sentences across seven ambiguity categories.
Using LaViSA, we conduct a comprehensive evaluation of diverse VLMs, including both proprietary and open-source models with varying parameter scales and reasoning capabilities. 
Experimental results show that although recent VLMs can leverage visual scenes to resolve structural ambiguity to a some extent, they still struggle with certain ambiguity types and visually subtle semantic distinctions, indicating remaining limitations in resolving structural ambiguity using visual scenes. 
\end{abstract}

%% file: contents/1_introduction.tex
\section{Introduction}

\begin{figure}[t]
    \centering
    \includegraphics[keepaspectratio, width=\linewidth]{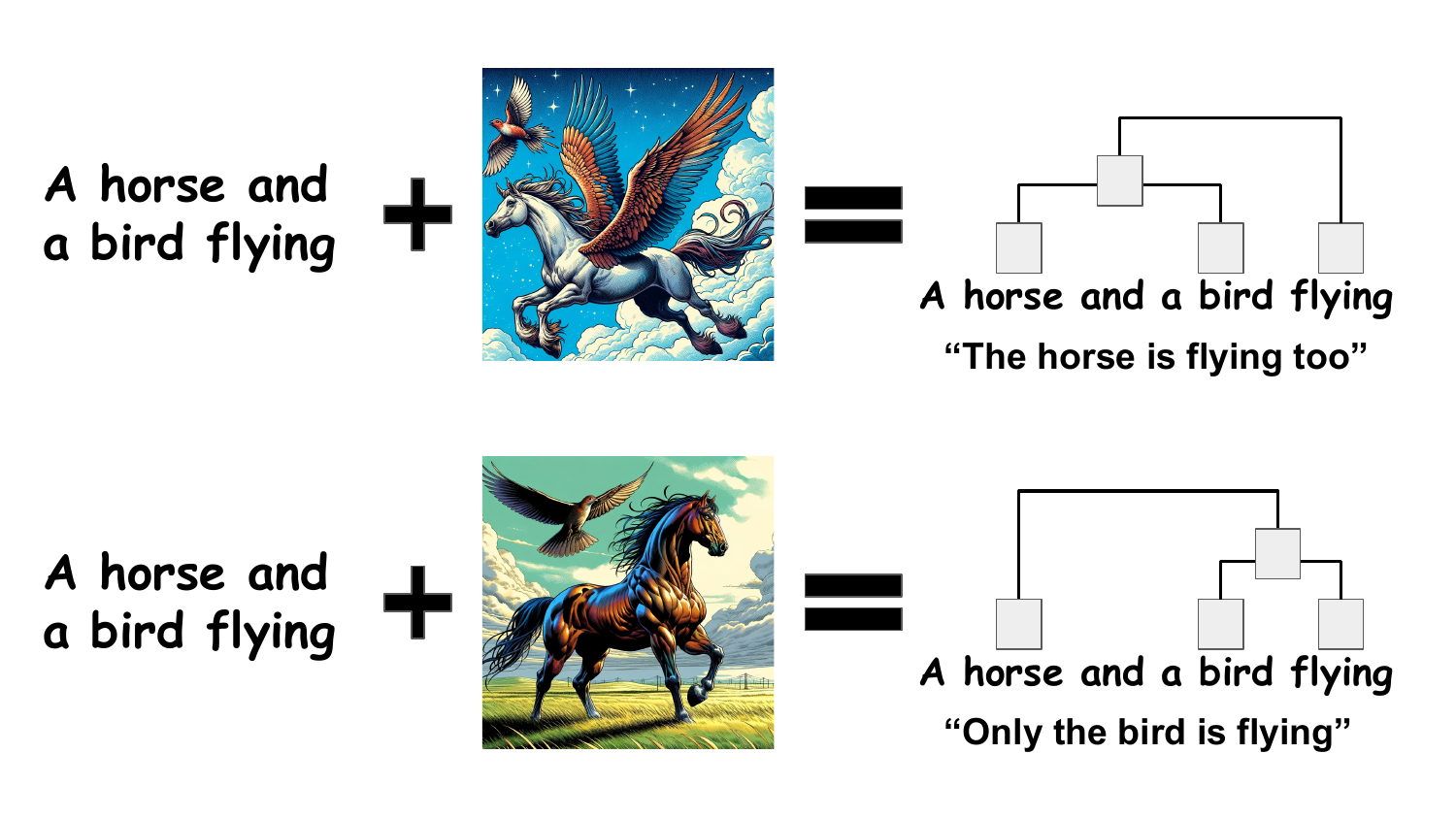}
    \caption{An example phrase of structural ambiguity \textit{A horse and a bird flying}. Given clarifying images, the phrase can be visually disambiguated into either the horse flying along with the bird or not.}
    \vspace{1mm}    
    \label{fig:one}
\end{figure}

Structural ambiguity arises when a sentence supports multiple interpretations due to its syntactic structure~\cite{Chomsky1965_aspects_of_syntax, hindle-rooth-1993-structural}. 
For example, in a phrase such as \textit{a horse and a bird flying} (Figure~\ref{fig:one}), the described situation differs depending on whether the modifier \textit{flying} is attached only to \textit{bird} or both the \textit{bird} and the \textit{horse}. 
Because sentence comprehension involves incrementally constructing syntactic interpretations, structural ambiguity can lead to difficulties in processing and interpreting diverse semantics, which are observed in both humans and Large Language Models~\cite{Fujita2021OnTP, Arehalli2022SyntacticSF, Amouyal2025WhenTL, Chandio}.


Structural ambiguity can be reduced by introspection~\cite{Chomsky1965_aspects_of_syntax}, rewriting an utterance into a less ambiguous form~\cite{duan-etal-2016-generating, stengel-eskin-etal-2023-chicken}. 
However, in real-world interaction, it is difficult to consistently apply introspection to every sentence encountered. 
In practice, problems arising from structural ambiguity are relatively rare in our daily lives because extra-sentential information, such as dialogue history, prosody, and visual scenes, plays an important role in real-world interaction~\cite{DeVault2009LearningTI, Kuribayashi2023DoesVA, Widiaputri2023SpeechRA}. 
Among these, visual scenes are frequently regarded as a particularly important cue in real-world interaction, with their high accessibility~\cite{Reiter2005ChoosingWI, Roy2005ConnectingLT, Coco2015TheIO, Hutmacher2019WhyIT}. 
In Figure~\ref{fig:one}, humans can use visual scenes to resolve an ambiguous phrase by narrowing down possible structural interpretations during sentence comprehension~\cite{Coco2015TheIO}.
We refer to this process as \textbf{Visual Disambiguation}.


As Vision and Language Models (VLMs) are expected to collaborate with humans in real-world interaction, measuring their human-like visual disambiguation capabilities is an important step towards practical applications. 
A large body of prior work has primarily been focused on compositionality, evaluating whether models can correctly ground sentences whose meanings differ due to changes in word order~\cite{Thrush2022WinogroundPV, Yuksekgonul2022WhenAW, Yamada2022WhenAL, Chung2025MASSOL}. 
However, ambiguity poses a distinct challenge beyond compositionality, as multiple meanings are tied to a single sentence.
In parallel, various types of ambiguity under multimodal settings have also been discussed~\cite{Berzak2015DoYS, Delecraz2018AddingSA, Mehrabi2023ResolvingAI, stengel-eskin-etal-2023-chicken, Chung2024CanVL, Kuribayashi2023DoesVA, Wang2025MUCARBM, Zhou2025FOCUSEP}.
Among these, this study focuses specifically on cases in which the interpretation of a sentence is uniquely determined by external information in the form of visual scenes. 
To address this problem, this paper makes two important contributions.
% ...

The first contribution is the proposal of \textbf{LaViSA} (\textbf{La}nguage and \textbf{Vi}sion \textbf{S}tructural \textbf{A}mbiguity), a benchmark designed to evaluate the ability to use visual scenes to determine a unique semantic interpretation for sentences with structural ambiguity associated with multiple possible interpretations. 
This dataset consists of structurally ambiguous sentences, their disambiguated interpretative sentences, and corresponding images, comprising a total of 1,503 samples across seven categories. 
For example, in Figure~\ref{fig:one}, \textit{a horse and a bird flying} is ambiguous. 
By associating it with interpretative sentences such as \textit{the horse is flying too} and \textit{only the bird is flying}, the ambiguity is explicitly resolved. 

The second contribution is a comprehensive study of the visual disambiguation capabilities of a wide range of existing VLMs using LaViSA.
In this evaluation, we assess their ability to select the correct interpretation corresponding to an image, given a pair consisting of a structurally ambiguous sentence and an image that resolves its ambiguity. 
This evaluation was conducted for both families of models: proprietary VLMs~\cite{OpenAI2025GPT5, google-2025-gemini3-1-pro, google-2025-gemini3-1-flash-lite} and open-source VLMs~\cite{An2025LLaVAOneVision15FO, Kamath2025Gemma3T, Bai2025Qwen3VLTR}.  

Our experimental results show that visual scenes provide effective additional information for resolving structural ambiguity, and that many VLMs can leverage such information successfully.
However, performance varies across ambiguity categories, and models often fail to distinguish the semantics of different visual scenes associated with the same ambiguous sentence.
Through a qualitative analysis of rationales generated by a reasoning-oriented VLM, we found indications that VLMs may struggle to fully grasp visual semantics and link them to the correct disambiguated interpretations.
These results highlight important challenges for future research on enabling VLMs to fully comprehend and integrate semantics across vision and language. We make our codes and data publicly available.\footnote{Code: \url{https://github.com/Lee-KocaTKa/LaViSA}\newline Dataset: \url{https://huggingface.co/datasets/Kerimu/LaViSA/tree/main}}

%% file: contents/2_related_work.tex
\section{Related Work}

\subsection{Compositionality in VLMs}
VLMs have been shown to struggle with aligning linguistic structure to visual scenes, particularly in settings that require sensitivity to fine-grained semantic composition~\cite{Tang2022WhenAL, ma-etal-2024-examination, Huang2024conme, Mitra2024CVPR}. 
Prior work has proposed benchmarks that probe compositional understanding by altering word order to induce meaning changes~\cite{Thrush2022WinogroundPV, Yuksekgonul2022WhenAW}.  
Our work extends this line of research by focusing on structural ambiguity, which requires models to handle multiple meanings of a single sentence, unlike compositionality, which concerns a single intended meaning. 





\subsection{Visual Disambiguation}

Prior work has introduced benchmarks that use visual scenes to evaluate linguistic ambiguity resolution~\cite{Chung2024CanVL,inadumi-etal-2024-gaze, Kuribayashi2023DoesVA, Wang2025MUCARBM, Nam_2025_ICCV}. 
However, these benchmarks do not systematically isolate different types of structural ambiguity or evaluate whether VLMs can resolve them using visual scenes. 

The Language and Visual Ambiguity (LAVA) corpus is one of the few datasets explicitly designed to address structural ambiguity using visual annotations~\cite{Berzak2015DoYS}. 
The Text-to-Image Ambiguity Benchmark (TAB) later extended LAVA with improved textual annotations and broader coverage for text-to-image generation~\cite{Mehrabi2023ResolvingAI}. 
However, despite their importance as early resources, LAVA remains limited in scale and annotation quality for evaluating modern VLMs~\cite{Mehrabi2023ResolvingAI, Yamaki2023HolographicCP}, while TAB lacks visual scenes and targets prompt disambiguation in text-to-image generation rather than visual disambiguation by VLMs. 


Recent work has extended LAVA using image and language generation models~\cite{BetkerImprovingIG, Achiam2023GPT4TR} to create a large-scale benchmark of sentences paired with controlled images for evaluating VLMs' underspecification reasoning with visual scenes~\cite{Zhou2025FOCUSEP}.
However, this benchmark focuses on underspecification and visual reasoning, whereas our goal is to evaluate whether VLMs can use visual scenes to resolve structural ambiguity. 
Building on a similar data-generation methodology, we construct LaViSA specifically for visual disambiguation and discuss the challenges VLMs face in this setting.

%% file: contents/3_LaViSA.tex
\section{LaViSA: \textbf{La}nguage and \textbf{Vi}sion \textbf{S}tructural \textbf{A}mbiguity Benchmark}
\label{3}
We constructed \textbf{La}nguage and \textbf{Vi}sion \textbf{S}tructural \textbf{A}mbiguity Benchmark (\textbf{LaViSA}), which consists of 1,503 samples across seven structural ambiguity categories.
Each sample in LaViSA consists of an ambiguous sentence, multiple disambiguated sentences, and images generated by a text-to-image model~\cite{BetkerImprovingIG}, which allows us to control the image correspondence to the disambiguated sentences.


\begin{figure}[t]
 \centering
 \includegraphics[width=\linewidth]{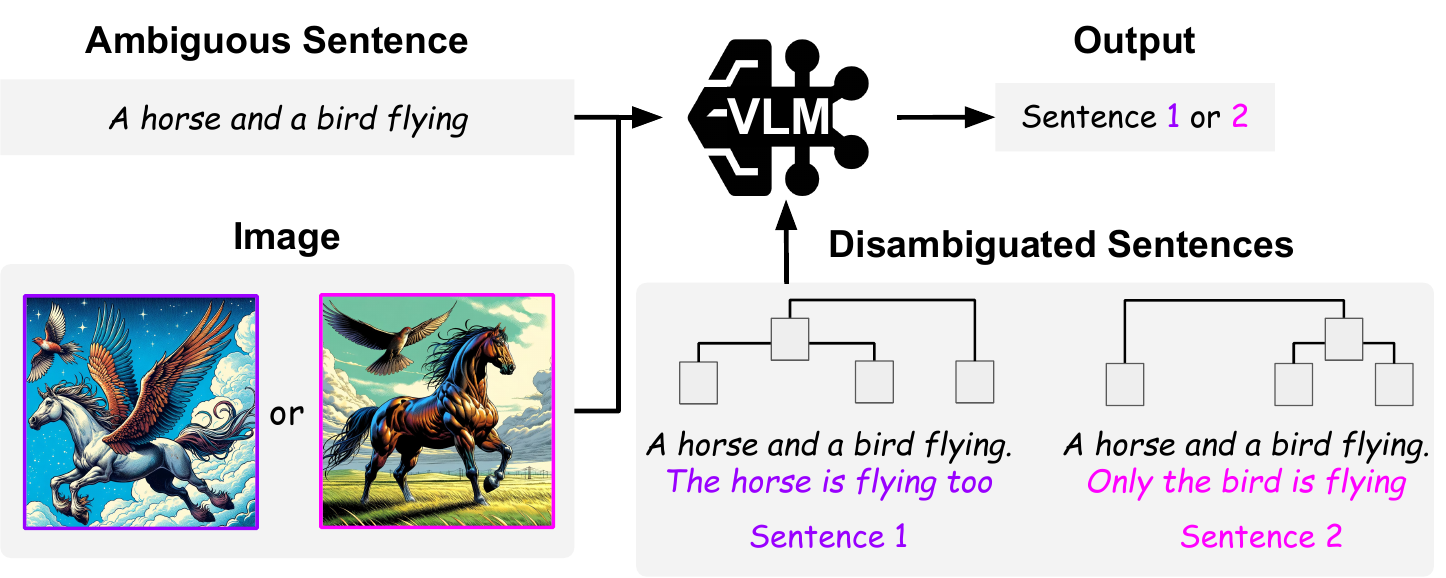}
 \caption{Overview of the visual disambiguation task}
 \label{fig:task}
\end{figure}

\subsection{Task Definition}
\label{3.1}

Given an ambiguous sentence, an image, and a set of interpretations represented by disambiguated sentences, visual disambiguation in LaViSA requires selecting an interpretation that corresponds to the image.
The VLM is evaluated on whether it can select the correct interpretation from the given inputs.
For example, Figure~\ref{fig:task} shows a case where the VLM must select an interpretation that matches the image from two syntactically distinct interpretations of the same ambiguous sentence.


A direct evaluation methodology for visual disambiguation would be to require VLMs to generate disambiguated sentences.
However, automatic evaluation metrics that compare generated sentences with references, such as BERTScore~\cite{Zhang2020BERTScoreET}, strongly depend on specific pre-trained language models.
To avoid this metric-dependence problem, we formulate visual disambiguation as a controlled multiple-choice task.


\begin{figure}[t]
  \centering
  \includegraphics[width=\columnwidth]{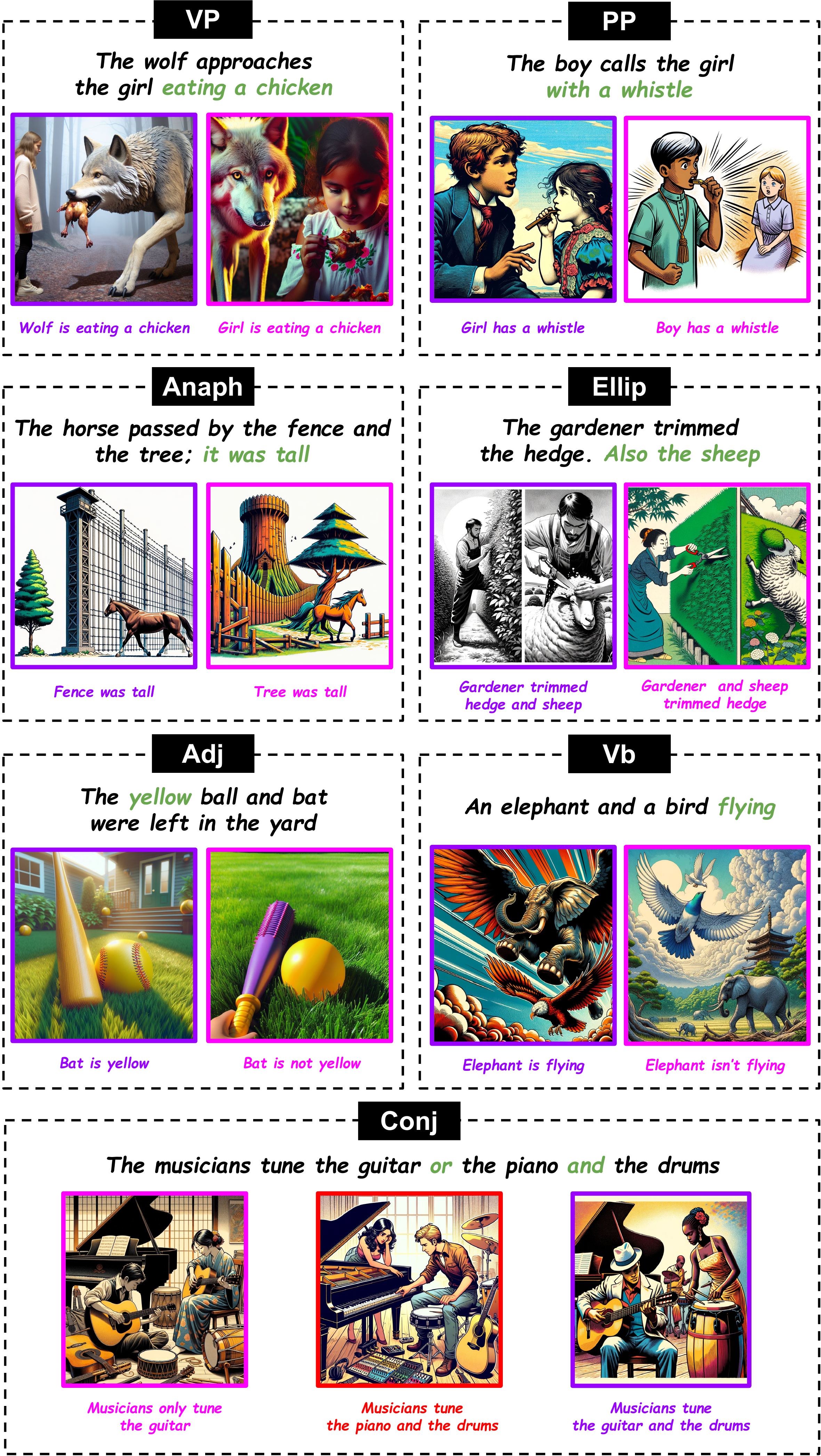}
  \caption{Examples from LaViSA across ambiguity categories. Each ambiguous sentence is paired with two or three disambiguated sentences and corresponding visual scenes.}
  \label{fig:categories}
\end{figure}

\subsection{Ambiguity Categories}
\label{3.2}

We categorise instances in LaViSA to enable a diversified evaluation.
Our categories are designed to distinguish major linguistic patterns of structural ambiguity that are plausibly resolvable using a visual scene. 
We build these categories upon the samples introduced in TAB~\cite{Mehrabi2023ResolvingAI}, which was originally designed for text-to-image generation and consists of ambiguous sentences.

Figure~\ref{fig:categories} illustrates the ambiguity categories in LaViSA, which are defined as follows:
\begin{description}[leftmargin=3.2em, labelwidth=2.8em, itemsep=1pt, parsep=1pt]
    \item[\textbf{VP}] Verb Phrase Attachment: Ambiguity arises when a verb phrase can attach to more than one part of the sentence.
    \item[\textbf{PP}] Preposition Phrase Attachment: A prepositional phrase can modify multiple possible heads.
    \item[\textbf{Anaph}] Anaphora: A pronoun or referring expression has more than one plausible antecedent.
    \item[\textbf{Ellip}] Ellipsis: An omitted phrase can be interpreted in multiple ways.
    \item[\textbf{Adj}] Adjective Scope: An adjective can modify either a single noun or an entire coordinated noun phrase.
    \item[\textbf{Vb}] Verb Scope: A verb-derived modifier may apply to one or more coordinated elements.
    \item[\textbf{Conj}] Conjunction Scope: Coordinating conjunctions (e.g., and, or) group sentence elements in more than one way.
\end{description}
TAB originally grouped adjective scope, verb scope, and conjunction scope into a single category.
We expand this category into three separate categories in LaViSA to distinguish different mechanisms of structural interpretation.
Detailed revisions from TAB are provided in Appendix~\ref{appen:A}.

\begin{figure}[t]
  \centering
  \includegraphics[width=\columnwidth]{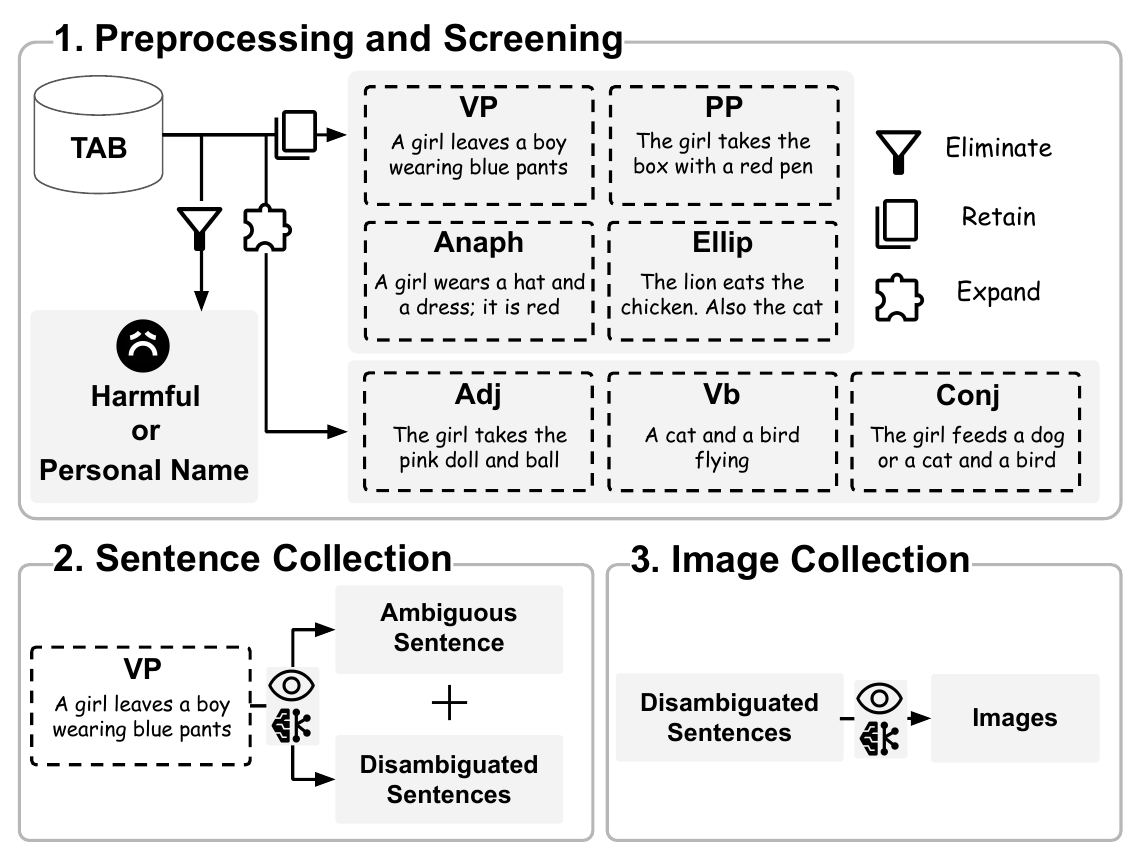}
  \caption{Data collection process of LaViSA. Each step involves manual inspection.} 
  \label{fig:data_construct}
\end{figure}

\subsection{Data Collection}
\label{3.3}

Figure~\ref{fig:data_construct} shows the construction process of LaViSA, where each step involves manual inspection.
Starting from TAB, we screen and preprocess seed data, and use a large language model~\cite{Achiam2023GPT4TR} to augment ambiguous sentences and their disambiguated sentences.
We then use a text-to-image model~\cite{BetkerImprovingIG} to collect images corresponding to the disambiguated sentences in a controlled manner.
Details of data collection, including the prompts, are provided in Appendix~\ref{appen:B}.

\subsubsection{Preprocessing and Screening}
Based on the ambiguity categories, we select ambiguous sentences and their disambiguated clauses from TAB as the seed data for LaViSA.
The text of TAB consists of ambiguous sentences with disambiguated clauses, which are combined to form complete disambiguated sentences.
Following the category definitions, we retain and expand TAB's original ambiguity categories from five to seven categories.
We then filter the seed data based on whether the sentences clearly demonstrate structural ambiguity and depict situations that can be visually presented.
We also eliminate samples that contain harmful content or personal names.

\subsubsection{Sentence Collection}
\label{3.3.1}
We input the seed data into GPT-4o\footnote{\url{gpt-4o-mini-2024-07-18}}~\cite{Achiam2023GPT4TR} to generate additional ambiguous sentences together with their disambiguated sentences in a similar format.
Each ambiguous sentence is paired with two or three disambiguated sentences.
This process results in 700 ambiguous sentences, with 100 sentences per category, and 1,503 disambiguated sentences in total.
During generation, we discarded or minimally revised outputs flagged by the model for harmful content or personal names.

\subsubsection{Image Collection}
\label{3.3.2}
For each disambiguated sentence, we generate a corresponding image using DALL-E 3\footnote{We used \url{dall-e-3} from Dec. 11, 2024 to Apr. 5, 2025.}~\cite{BetkerImprovingIG}.
To test for sensitivity to image style, approximately half of the images are generated in a cartoon style and the rest in a photorealistic style.

\begin{table}[t]
    \centering
    \scalebox{0.78}{
    \begin{tabular}{ccccccc|c}
    \toprule
    \textbf{VP} & \textbf{PP} & \textbf{Anaph} & \textbf{Ellip} & \textbf{Adj} & \textbf{Vb} & \textbf{Conj} & \textbf{All} \\
    \cmidrule{1-8}
    96.5 & 92.5 & 94.5 & 89.6 & 89.5 & 94.5 & 93.3 & 92.9 \\
    \bottomrule
    \end{tabular}
}
\caption{Human validation of generated images against intended disambiguated sentences in LaViSA.}
\label{table:human_eval}
\end{table}

To validate the quality of the generated images, we conduct an independent human validation to assess whether each image corresponds to its intended disambiguated sentence.
Two annotators who were not involved in the data construction process are given an image together with candidate disambiguated sentences for all samples, without the ambiguous sentence, and asked to select the interpretation best supported by the image.
As shown in Table~\ref{table:human_eval}, the validation accuracy exceeds 90\% overall, while even the lowest-performing ambiguity category remains close to 90\%.
These results suggest that the generated images generally correspond to the intended disambiguated sentences and are interpretable for human readers, supporting the validity of LaViSA as a benchmark for visual disambiguation.

%% file: contents/4_experiment.tex
\begin{table}[t]
    \centering
    \scalebox{0.9}{
    \begin{tabular}{lcc}
    \toprule
    \textbf{Models} & \textbf{PT-Acc} & \textbf{PS-Acc} \\
    \midrule
    \multicolumn{3}{c}{\textbf{\textit{Proprietary VLMs}}} \\
    GPT-5.2 & 83.4 & 67.1 \\
    Gemini 3.1 Pro & \textbf{88.9} & \textbf{77.7} \\
    Gemini 3.1 Flash-Lite & 84.0 & 69.0 \\
    \cmidrule(lr){1-3}
    \multicolumn{3}{c}{\textbf{\textit{Open-Source VLMs (Instruct)}}} \\
    LLaVA-One-8B & 73.4 & 49.0 \\
    Gemma3-4B & 51.0 & 9.1 \\
    Gemma3-12B & 69.9 & 42.3 \\
    Gemma3-27B & 71.7 & 43.9 \\
    Qwen3-VL-4B & 78.2 & 57.0 \\
    Qwen3-VL-8B & 72.5 & 49.4 \\
    Qwen3-VL-32B & \underline{81.9} & \underline{64.3} \\
    \cmidrule(lr){1-3}
    \multicolumn{3}{c}{\textbf{\textit{Open-Source VLMs (Thinking)}}} \\
    Qwen3-VL-4B & 65.9 & 44.0 \\
    Qwen3-VL-8B & 71.1 & 50.3 \\
    Qwen3-VL-32B &  78.9 & 61.1 \\
    \bottomrule
    \end{tabular}
}
\caption{Evaluation results on LaViSA, averaged over ambiguity categories. Bold scores indicate the best overall performance for each metric, and underlined scores indicate the best performance among open-source VLMs.}
\label{table:main_all}
\end{table}

\begin{table}[t]
    \centering
    \scalebox{0.9}{
    \begin{tabular}{lcc}
    \toprule
    \textbf{Image Styles} & \textbf{PT-Acc} & \textbf{PS-Acc} \\
    \midrule
    Cartoon & 83.4 & 67.3 \\
    Photo & 92.2 & 84.3 \\
    All & 88.9 & 77.7 \\
    \bottomrule
    \end{tabular}
}
\caption{Ablation study on input image styles using Gemini 3.1 Pro, the best-performing model in Table~\ref{table:main_all}. }
\label{table:ablation_style}
\end{table}

\begin{table*}[t]
\centering
\scalebox{0.6}{
    \begin{tabular}{l cc cc cc cc cc cc cc}
    \toprule
    & \multicolumn{2}{c}{\textbf{VP}} & \multicolumn{2}{c}{\textbf{PP}} & \multicolumn{2}{c}{\textbf{Anaph}} & \multicolumn{2}{c}{\textbf{Ellip}} & \multicolumn{2}{c}{\textbf{Adj}} & \multicolumn{2}{c}{\textbf{Vb}} & \multicolumn{2}{c}{\textbf{Conj}} \\
    
    \cmidrule(lr){2-3}
    \cmidrule(lr){4-5}
    \cmidrule(lr){6-7}
    \cmidrule(lr){8-9}
    \cmidrule(lr){10-11}
    \cmidrule(lr){12-13}
    \cmidrule(lr){14-15}
    
    \textbf{Model} & PT-Acc & PS-Acc & PT-Acc & PS-Acc & PT-Acc & PS-Acc & PT-Acc & PS-Acc & PT-Acc & PS-Acc & PT-Acc & PS-Acc & PT-Acc & PS-Acc \\
    \midrule
    \multicolumn{15}{c}{\textbf{\textit{Proprietary Models}}} \\
    GPT-5.2 
    & 95.0 & 90.0 
    & \textbf{94.0} & \textbf{89.0} 
    & \textbf{90.0} & \textbf{81.0} 
    & 71.1 & 37.0 
    & 87.5 & 76.0 
    & 82.5 & 66.0 
    & 73.0 & 31.0 
    \\
    Gemini 3.1 Pro 
    & \textbf{97.0} & \textbf{94.0} 
    & \textbf{94.0} & 88.0 
    & \textbf{90.0} & 80.0 
    & \textbf{83.2} & \textbf{67.0} 
    & \textbf{90.5} & \textbf{82.0} 
    & \textbf{84.5} & \textbf{73.0} 
    & \textbf{85.0} & \textbf{60.0} \\
    
    Gemini 3.1 Flash-Lite
    & 93.5 & 87.0 
    & 94.5 & 89.0 
    & 82.6 & 69.0 
    & 70.8 & 42.0 
    & 90.5 & 82.0 
    & 84.0 & 70.0 
    & 76.3 & 44.0 
    \\

    \cmidrule(lr){1-15}
    \multicolumn{15}{c}{\textbf{\textit{Open-Source VLMs (Instruct)}}} \\
    
    LLaVA-One-8B
    & 74.0 & 48.0
    & 87.5&  75.0
    & 79.1 & 59.0 
    & 58.4&  18.0
    & 84.0 & 68.0
    & 78.0&  57.0
    & 62.7 & 26.0
    \\
    
    Gemma3-4B
    &52.5 & 5.0
    & 60.5& 23.0
    &52.7 & 8.0
    &54.5 & 12.0
    & 54.0& 11.0
    &51.5 & 5.0
    & 38.0& 0.0
    \\
    
    Gemma3-12B
    &67.5 & 36.0
    & 78.5& 58.0
    &75.1 & 52.0
    &61.9 &  25.0
    & 79.0& 61.0
    & 72.5& 46.0
    & 59.7& 18.0
    \\
    
    Gemma3-27B
    & 65.0 & 30.0
    & 85.0 & 70.0
    & 76.1 & 54.0
    & 61.4 & 24.0
    & 76.0 & 56.0
    & 73.0 & 51.0
    & 67.3 & 22.0
    \\

    Qwen3-VL-4B
    &82.5 & 65.0
    &88.5 & 78.0
    &81.6 & 64.0 
    &61.4 & 25.0
    & 82.0 & 65.0
    & 80.0&  62.0
    &73.7 & 40.0
    \\
    
    Qwen3-VL-8B
    &71.0 & 42.0
    &83.0 & 67.0
    &80.1 & 64.0
    & \underline{66.8} & \underline{35.0}
    & 85.0 & 71.0
    & 79.0 & 60.0
    & 51.7 & 7.0
    \\
    
    Qwen3-VL-32B
    &86.0 & 72.0
    & \underline{93.5} & \underline{88.0}
    & \underline{84.6} & \underline{71.0}
    & 63.9& 30.0
    & 87.0 & 75.0
    & \underline{82.0} & \underline{66.0}
    & \underline{78.3} & \underline{48.0}
    \\
    \cmidrule(lr){1-15}
    \multicolumn{15}{c}{\textbf{\textit{Open-Source VLMs (Thinking)}}} \\
    Qwen3-VL-4B
    & 79.0 & 63.0
    & 81.0 & 64.0
    & 62.2 & 38.0 
    & 51.0 & 15.0
    & 82.0 & 67.0 
    & 74.0 & 55.0
    & 43.3 & 6.0
    \\
    
    Qwen3-VL-8B
    & 86.0 & 74.0
    & 83.5 & 69.0
    &  74.1 & 52.0  
    & 59.4 & 27.0
    & 80.5 & 66.0
    & 80.0 & 61.0
    & 46.3 & 3.0 
    \\
    
    Qwen3-VL-32B
    & \underline{92.5} & \underline{85.0}
    & 87.5 & 76.0
    & \underline{84.6} & \underline{71.0}
    & 65.8 & \underline{35.0}
    & \underline{87.5} & \underline{76.0}
    & 80.5 & 65.0
    & 62.3 & 20.0
    \\
    
    \cmidrule(lr){1-15}
    
    Random Chance
    & 50.0 & 25.0
    & 50.0 & 25.0
    & 50.0 & 25.0
    & 50.0 & 25.0
    & 50.0 & 25.0
    & 50.0 & 25.0
    & 33.3 & 11.1
    \\    
    \bottomrule
    \end{tabular}
}

\caption{Evaluation results on LaViSA for each ambiguity category. Bold and underlined scores follows Table~\ref{table:main_all}.}
\label{table:main_per_category}
\end{table*}

\section{Experiment}
\label{4}

We quantitatively clarify existing VLMs' ability to handle structural ambiguity using benchmark scores on the task defined in LaViSA. We first clarify the experimental setup and then show results. As discussed in Section~\ref{3.1}, the main evaluation task is formulated as visual disambiguation. In each trial, a model is given an ambiguous sentence and a clarifying image, and is required to select the interpretation best supported by the image from two or three candidate interpretations derived from the same ambiguous sentence.

\subsection{Experimental Setup}

\label{4.1}


We report two evaluation metrics corresponding to the visual disambiguation: \textbf{Per-trial Accuracy} (PT-Acc) and \textbf{Per-sentence Accuracy} (PS-Acc).
PT-Acc is the percentage of individual trials in which the model selects the correct interpretation. 
PS-Acc is the percentage of ambiguous sentences for which the model correctly resolves ambiguity with all associated interpretations. 
For example, if an ambiguous sentence has two corresponding images, the model must select the correct interpretation for both images in order for the ambiguous sentence to be counted as correct. This metric evaluates whether the model consistently changes its prediction when the clarifying image changes, rather than always preferring the same candidate interpretation.



We evaluate both proprietary and open-source VLMs. 
For proprietary models, we include GPT-5.2~\cite{OpenAI2025GPT5}, Gemini 3.1 Pro~\cite{google-2025-gemini3-1-pro}, and Gemini 3.1 Flash-Lite~\cite{google-2025-gemini3-1-flash-lite}. 
Because LaViSA is constructed with assistance from GPT-family models, evaluating only GPT-family models could make the results dependent on a single model family. 
We therefore include Gemini-family models to examine whether the observed trends generalise across independently developed proprietary VLMs. For open-source models, we evaluate the instruction-tuned versions of LLaVA-OneVision-1.5~\cite{An2025LLaVAOneVision15FO} (LLaVA-One), Qwen3-VL~\cite{Bai2025Qwen3VLTR}, and Gemma3~\cite{Kamath2025Gemma3T}. For Qwen3-VL and Gemma3, we evaluate multiple parameter-scale variants, resulting in six model variants in total. In addition, for Qwen3-VL, we compare both standard instruction-tuned variants and reasoning-oriented (“thinking”) variants within the same model family.

\subsection{Results}
\label{4.3}

Table~\ref{table:main_all} reports the main experimental results across the entire LaViSA benchmark. Table~\ref{table:ablation_style} further breaks down the results of Gemini 3.1 Pro model according to image style, while Table~\ref{table:main_per_category} elaborates them by ambiguity categories. 

\subsubsection{Overall Performance}
\label{4.3.1}
\paragraph{Proprietary Models}
Table~\ref{table:main_all} shows that proprietary models generally outperform open-source models on both PT-Acc and PS-Acc. 
Among all evaluated models, Gemini 3.1 Pro achieves the strongest overall performance, obtaining the highest scores on both metrics. 
GPT-5.2 also demonstrates strong performance, while Gemini 3.1 Flash-Lite outperforms GPT 5.2 despite its lightweight configuration.   

\paragraph{Open-Source Models}
Among open-source models, Qwen3-VL variants consistently outperform other model families. 
In particular, Qwen3-VL-32B achieves the strongest overall performance, becoming the only open-source model to surpass 80\% PT-Acc. 
Within the Qwen3-VL family, however, reasoning-oriented (“thinking”) variants generally underperform their instruction-tuned counterparts. 


\paragraph{Model Scale}

In general, the best performance within each model family is achieved by the largest variant. 
However, the overall results do not exhibit a simple monotonic relationship between model size and performance. For example, the largest 27B model in the Gemma3 family is outperformed by smaller models from other families. Moreover, Qwen-VL-4B-Instruct outperforms its 8B counterpart, while the 32B model approaches or exceeds much larger proprietary models in some categories, as shown in Table~\ref{table:main_per_category}. 

\subsubsection{Conditional Performance}
\label{4.3.2}
\paragraph{By Image Style}

Table~\ref{table:ablation_style} reports experimental results based on image styles using Gemini 3.1 Pro, the best-performing model in Table~\ref{table:main_all}. 
Performance on photorealistic-style images is higher than on cartoon-style images in both PT-Acc and PS-Acc. One possible reason is that modern VLMs are more strongly aligned with natural-image distributions commonly observed during pre-training. Nevertheless, the strong performance on photorealistic-style images suggests that LaViSA does not rely solely on artifacts specific to synthetically generated imagery. Rather, the results indicate that visual disambiguation performance remains meaningful across different visual styles.  

\paragraph{By Ambiguity Category}
The experimental results on Table~\ref{table:main_per_category} reveal differences across ambiguity categories. Attachment-related categories such as VP and PP yield the highest accuracies across most models. Scope-related categories, especially Adj and Vb, are also relatively easier than the remaining categories. By contrast, Ellip and Conj remain consistently difficult across models. One possible explanation is that attachment and scope ambiguities are often grounded in relatively standardised visual scenes, such as the position, holder, or attribute of a particular object entity. In contrast, Ellip and Conj frequently require models to recover more global semantic structures supplemented by zero anaphora and related phenomena, or grouping relations spanning multiple entities and events. These categories may therefore require deeper semantic understanding beyond detecting salient visual entities or attributes. We further investigate these difficulties through qualitative analysis in Section~\ref{5}. 

\begin{figure*}[t]
    \centering
    \includegraphics[width=\textwidth]{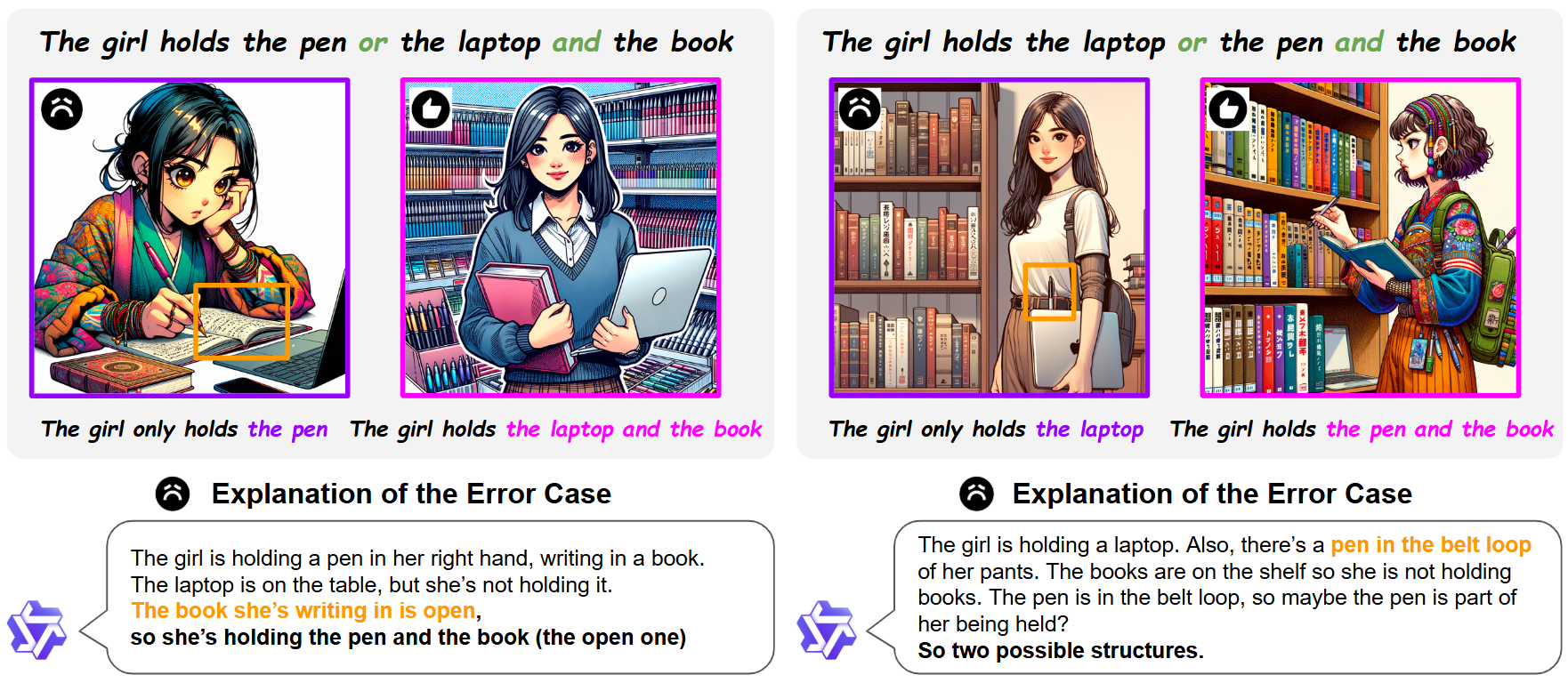}
    \caption{
    Error cases involving visually salient but structurally misleading objects.
    Pink annotations indicate the correct interpretation, while purple annotations indicate incorrect interpretation. 
    Orange annotations indicate wrong explanations and visual cues from the VLM's explanation.}
    \label{fig:error_cases_one}
\end{figure*}

\subsubsection{Visual Scenes as Additional Information}

\paragraph{Effectiveness}

The random chance baseline in Table~\ref{table:main_per_category} represents the expected accuracy when selecting among candidate interpretations without access to clarifying visual information. As shown in Table~\ref{table:main_per_category}, most evaluated models consistently outperform this baseline across ambiguity categories, with the strongest models achieving accuracies exceeding 90\% in several cases. These results suggest that visual scenes indeed provide useful additional information for resolving structural ambiguity, and that current VLMs are capable of leveraging such information to a substantial extent.

\paragraph{Concerns}
There is generally a relationship between PT-Acc and PS-Acc such that PS-Acc is close to the square or cube of PT-Acc. Since the combinations used in PS-Acc are mutually related, this scope gap is expected to be smaller if interpretations are properly associated for a given instance. In other words, this suggests that current VLMs do not consistently track how changes in visual evidence correspond to different semantic structures. 


\begin{figure*}[t]
    \centering
    \includegraphics[width=\textwidth]{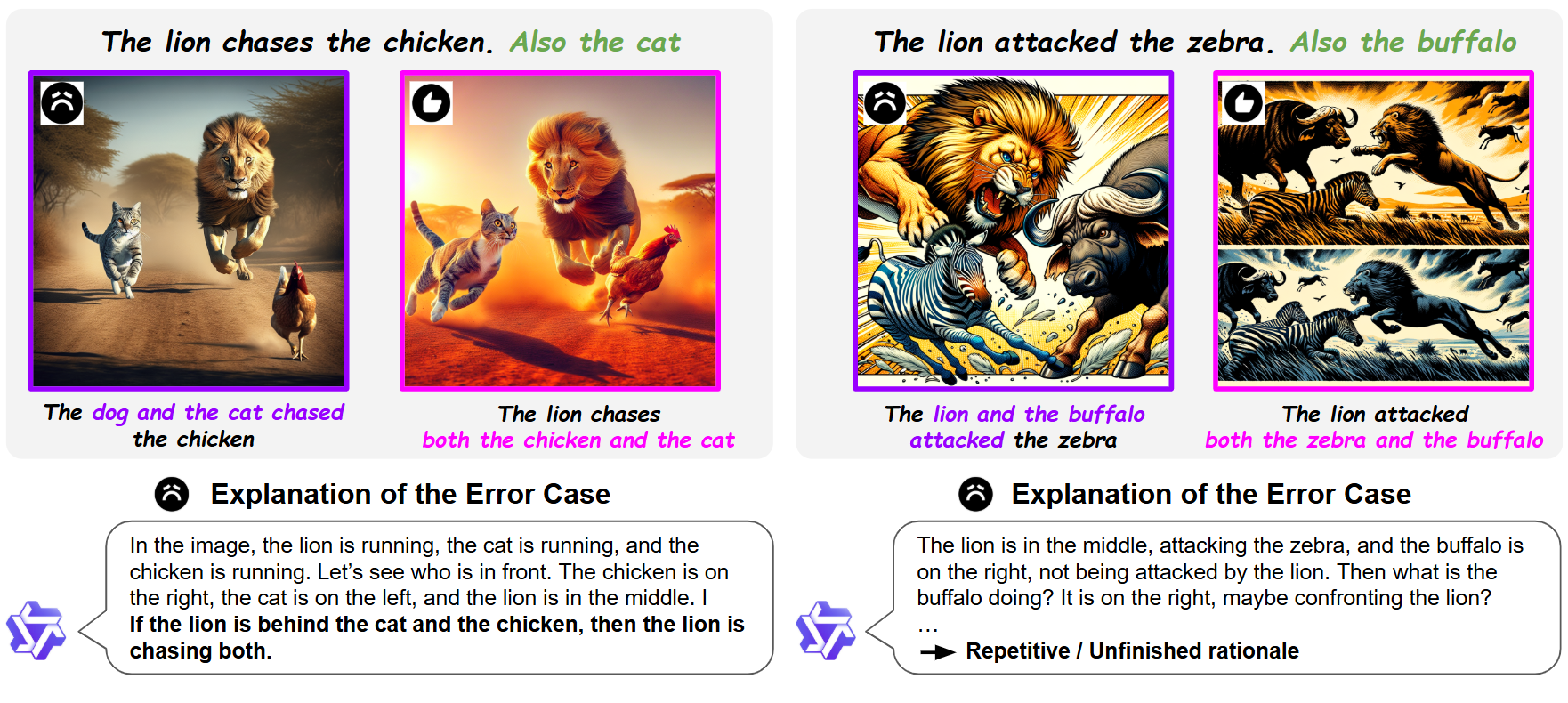}
    \caption{
    Error cases from the Ellip category. 
    The definition of pink and purple follows that of Figure~\ref{fig:error_cases_one}.
    }
    \label{fig:error_cases_two}
\end{figure*}

\section{Qualitative Analysis}
\label{5}

In this section, we conduct a qualitative analysis of cases in which current VLMs struggle with visual disambiguation. Our analysis is primarily based on rationales generated by reasoning-oriented variants of the Qwen3-VL Thinking family, with a particular focus on Qwen3-VL-32B-Thinking, the strongest performing model among them. As discussed in Section~\ref{4.3.2}, VLMs generally exhibit weaker performance on Ellip and Conj compared to other ambiguity categories. We hypothesise that the linguistic characteristics of these categories contribute to the observed difficulties and therefore analyse them separately in the following subsections. 

\subsection{Error Cases from Conjunction Scope}
\label{5.1}

\paragraph{Linguistic Characteristics}
Ambiguous sentences in the Conj category of LaViSA generally take the form of ``Subject Verb Object$_1$ or Object$_2$ or Object$_3$.'' Although all candidate objects are visually present, the semantic interpretation depends on the scope of coordination and how the objects are associated with the predicate structure.

\paragraph{Visually Present but Predicate-Unrelated Objects}

Figure~\ref{fig:error_cases_one} illustrates two cases from the Conj category derived from the same ambiguous sentence. The ambiguity depends on which objects are interpreted as arguments of the predicate \textit{hold}. In the successful case, the model correctly identifies that the girl holds the laptop and the book despite the presence of pens in the scene. In the failure case, however, the model incorrectly concludes that the girl also holds the book, although it is only associated with the writing activity.


In the second pair, the ambiguity arises in \textit{The girl holds the laptop or the pen and the book.} In the successful case, the model correctly recognises that the girl holds the pen and the book, while the laptop is merely placed on the shelf. In the failure case, the pen tucked into the girl's clothing appears to mislead the model. Although the pen is not being held in the relevant sense of the predicate \textit{hold}, the model still treats its visual presence as evidence for an alternative interpretation. 

These examples suggest that VLMs may struggle to identify the correct predicate--argument structure when visually salient but predicate-unrelated objects appear near the subject. 

\subsection{Error Cases from Ellipsis}
\label{5.2}

\paragraph{Linguistic Characteristics}
Ambiguous sentences in the Ellip category generally take the form of ``Subject Verb Object, Also Noun,'' where the ``Noun'' may either be another subject or object of the predicate verb. As a result, resolving this ambiguity requires VLMs to infer the underlying predicate--argument structure from the visual scene.

\paragraph{When Object Becomes the Subject}
Figure~\ref{fig:error_cases_two} illustrates failure cases from the Ellip category. In the first example, the model correctly recognises the lion as the sole chaser in the successful case, but fails to infer that the cat also chases the chicken in the contrasted interpretation. The rationale suggests that the model relies heavily on spatial configuration, treating the cat as the object simply it is a step ahead of the lion. 

In the second example, the ambiguity concerns whether the buffalo is another attacker or an additional patient of the attack. Although the model correctly recognises that the buffalo is not being attacked by the lion, it fails to determine whether the buffalo itself attacks the zebra, eventually falling into a repetitive reasoning loop. 

These examples suggest that it is difficult for VLMs to consistently infer subject--object relations from complex visual interactions, especially when semantic roles change across minimally contrasted scenes.

%% file: contents/5_conclusion.tex
\section{Conclusion}

We introduced LaViSA, a benchmark for evaluating whether VLMs can resolve structural ambiguity using visual scenes. 
LaViSA consists of ambiguous sentences, disambiguated sentences representing their interpretations, and corresponding images across seven ambiguity categories. 
Experimental results across a diverse set of VLMs suggest that visual scenes provide effective additional information for resolving structural ambiguity, and that current models can leverage such information to a certain extent. 
However, VLMs still exhibit limitations, particularly in ambiguity categories such as Conj and Ellip. Moreover, the gap between per-trial and per-sentence accuracy indicates that models struggle to consistently track semantic differences when the visual scene changes across minimally contrasted interpretations. 
Through qualitative analysis of model-generated rationales, we further observe that VLMs often fail to correctly integrate visual objects into predicate--argument structures and semantic relations. 
These findings highlight important challenges for structurally grounded visual understanding and suggest future directions for improving semantic grounding in VLMs. 

%% file: contents/z_appendix.tex
\section{Revision of Ambiguity Categories}
\label{appen:A}

The TAB dataset has defined the following seven categories of ambiguity~\cite{Mehrabi2023ResolvingAI}:
\begin{itemize}
    \item \textbf{Anaphora}: Ambiguity arises when pronouns or similar expressions refer to a previously mentioned entity, but there are multiple possible referents. (e.g. \textit{The girl looks at the bird and the butterfly; it is red}.)
    \item \textbf{Ellipsis}: Ambiguity caused by omitted elements in a sentence, resulting in multiple possible interpretations. (e.g. \textit{The lion eats the chicken. Also the cat}.)
    \item \textbf{Fairness}: Ambiguity occurs when the caption lacks specific attributes of an object, resulting in multiple possible visual interpretations. (e.g. \textit{The man dusting the floor}.)
    \item \textbf{Syntax-PP}: Ambiguity arises when it's unclear which part of the sentence a prepositional phrase is modifying. (e.g. \textit{The woman approached the chair with a bag}.)
    \item \textbf{Syntax-VP}: Ambiguity arises when it's unclear which part of the sentence a verb phrase is modifying (e.g. \textit{The man looked at a boy talking to a telephone}.)
    \item \textbf{Conjunction}: Ambiguity caused by the scope of verbs or adjectives connected to multiple nouns via conjunctions like \textit{and} or \textit{or}. (e.g. \textit{The girl holds the green chair and bag}.)
    \item \textbf{Miscellaneous}: A collective set of ambiguities which are not affiliated to either of the above six, but not quantitatively enough to form a separate category. (e.g. \textit{The chicken is ready to eat}.)
\end{itemize}

Among the original ambiguity types, Fairness was found to be exclusively related to image generation and was therefore unsuitable for our research, which focuses on semantic diversity rather than visual representation. Additionally, the Miscellaneous category contained too few instances. As a result, we excluded both of these categories from our study. Furthermore, we redefined the Conjunction category as a scope ambiguity problem and subdivided it into three finer-grained types: adjective scope, verb scope, and conjunction scope.

\section{Data Collection Details}
\label{appen:B}
\subsection{Data Screening and Modification}
While TAB was originally designed for image generation tasks, some of its samples included inappropriate content that was rejected by the generation model. 
One common issue involved violent verbs, such as kill, threaten, or hit (e.g., \textit{The girl killed the boy with a gun.}). 
Another issue was the inclusion of real-world political figures from the contemporary era, which also triggered rejection (e.g., \textit{Biden sits next to a girl worshipping Trump}).
To address these issues, we made the following modifications: violent verbs were replaced with neutral alternatives (e.g., \textit{greet}), and named political figures were replaced with descriptive phrases (e.g., \textit{the old man and the blonde man}) to preserve the intended ambiguity while avoiding rejection by the model. 

\subsection{Model Prompts}
For data collection, we used the following prompts in Table~\ref{tab:prompt-templates}. For caption augmentation, previous samples from TAB were given to the generation model to grant it a sense of the sentences it was supposed to create. \texttt{\{AMB\_TYPE\}} was formatted with the name and a description of the ambiguity type as follows:
\begin{itemize}
    \item \textbf{VP}: VP Attachment Ambiguity, occurring when it is unclear which part of a sentence a verb phrase is intended to modify
    \item \textbf{PP}: PP Attachment Ambiguity, occurring when it is unclear which part of a sentence a prepositional phrase is intended to modify
    \item \textbf{Anaph}: Anaphoric Ambiguity, which occurs when it is unclear which antecedent a particular anaphor refers to within a given context
    \item \textbf{Ellip}: Ellipsis Ambiguity, involving the omission of words or phrases that are understood from the context
    \item \textbf{Adj}: Adjective Scope Ambiguity, occurring when it is unclear how far the influence of an adjective extends within a sentence
    \item \textbf{Vb}: Verb Scope Ambiguity, occurring when it is unclear how far the influence of a verb extends within a sentence
    \item \textbf{Conj}: Conjunction Scope Ambiguity, occurring when it is unclear how far the influence of a conjunction coordinate, such as AND/OR extends within a sentence
\end{itemize}
Image generation prompts were carefully made to have the same semantic structure as that of the texts used for the experiments. For image styles, \textit{coloured cartoon} and \textit{coloured photography} were used.

\section{Experimental Details}
\label{appen:C}
Table~\ref{tab:model_cards} summarizes the VLMs used in our experiments together with their HuggingFace IDs or API names~\cite{OpenAI2025GPT5, google-2025-gemini3-1-pro, google-2025-gemini3-1-flash-lite, An2025LLaVAOneVision15FO, Bai2025Qwen3VLTR, Kamath2025Gemma3T}. 
Table~\ref{tab:experimentalprompts} shows the prompt template used for the evaluation experiments.
Open-source VLM experiments were conducted on a single NVIDIA RTX A6000 GPU, and inference on LaViSA took at most two days for each VLM.

\begin{table*}[p]
    \centering
    \scalebox{0.8}{
    \begin{tabular}{ll}
    \toprule
    \textbf{VLMs} & \textbf{HuggingFace ID or API Name} \\
    \midrule
    GPT-5.2& \url{OpenAI/gpt-5.2-2025-12-11} \\
    Gemini 3.1 Pro & \url{Gemini/gemini-3.1-pro-preview} \\
    Gemini 3.1 Flash-Lite & \url{Gemini/gemini-3-1-flash-lite} \\
    LLaVA-OneVision-1.5-8B-Instruct & \url{lmms-lab/LLaVA-OneVision-1.5-8B-Instruct} \\
    Qwen3-VL-4B-Instruct & \url{Qwen/Qwen3-VL-4B-Instruct} \\
    Qwen3-VL-8B-Instruct & \url{Qwen/Qwen3-VL-8B-Instruct} \\
    Qwen3-VL-32B-Instruct & \url{Qwen/Qwen3-VL-32B-Instruct} \\
    Qwen3-VL-32B-Thinking & \url{Qwen/Qwen3-VL-32B-Thinking} \\
    Gemma3-4b-it & \url{google/gemma-3-4b-it} \\
    Gemma3-12b-it & \url{google/gemma-3-12b-it} \\
    Gemma3-27b-it & \url{google/gemma-3-27b-it} \\
    \bottomrule
    \end{tabular}
}
\caption{Details of the VLMs for the experiments.}
\label{tab:model_cards}
\end{table*}

\begin{table*}[p]
\centering
\begin{tcolorbox}[fontupper=\scriptsize, title=\small Prompt for Caption Augmentation]
\begin{verbatim}
Hi, I’m making a dataset by extending the following examples. 
Output sentences in the following format: 
    - An ambiguous sentence having 2 or 3 possible meanings: 
        Avoid repeating common phrases and use a wide range of vocabulary and creative expression, 
        a variety of synonyms and idioms. 
    - Disambiguated sentences corresponded to an ambiguous sentence: 
        Do not say something else, but just 2 or 3 sentences. These sentences are connected slash. 
    - If I’m not satisfied, I will give you feedback. If I say good, then generate another round.
    - Create a text filled with detail that allows one to easily visualise the scene. 
The topic is {AMB_TYPE}. From now on, I will show you some of the examples. 
Example: {EXAMPLE_FROM_TAB}
\end{verbatim}
\end{tcolorbox}
\begin{tcolorbox}[fontupper=\scriptsize, title=\small Prompt for Image Generation]
\begin{verbatim}
Follow the given caption prompt and visual style to generate a faithful image. 
    Prompt: {CAPTION}
    Style: {coloured cartoon OR coloured photography}
\end{verbatim}
\end{tcolorbox}
\caption{Prompt templates used for data collection.}
\label{tab:prompt-templates}
\end{table*}

\begin{table*}[p]
\begin{tcolorbox}[fontupper=\scriptsize, title=\small Prompt for Experiments]
\begin{verbatim}
You are a vision-language model performing structural disambiguation.
You will be given an ambiguous caption and an image.
Your task is to select the correct interpretation of the caption based on the image.
Important:
    - All options are derived from the same ambiguous caption and differ only in structural interpretation.
    - You must use the image to decide.
    - Do not rely only on textual plausibility.
Caption: {CAPTION}
Question: Among the following options, which one correctly reflects the meaning of the caption given the image?

Please provide your answer in the following format:
Answer: 〈one of the options〉 as a single number (1, 2, 3, etc.)
\end{verbatim}
\end{tcolorbox}
\caption{Prompt template used for experiments.}
\label{tab:experimentalprompts}
\end{table*}